\documentclass[runningheads]{llncs}
\usepackage[T1]{fontenc}
\usepackage{graphicx}
\usepackage{booktabs}
\usepackage[misc]{ifsym}
\newcommand{\corr}{(\Letter)}

\usepackage{amsmath}
\usepackage{subcaption}
\usepackage{xcolor}
\usepackage{amssymb}
\usepackage{hyperref}
\usepackage{cleveref}
\begin{document}
\title{Enhancing interpretability of rule-based classifiers through feature graphs}
%
%
\author{
Christel Sirocchi\inst{1}\corr \and
Damiano Verda\inst{2}
}
\authorrunning{C. Sirocchi \& D. Verda}
%
\institute{Department of Pure and Applied Sciences, University of Urbino, \\ Piazza della Repubblica 13, 61029, Urbino, Italy  \\ \email{c.sirocchi2@campus.uniurb.it} \and
Rulex Innovation Labs, Via Felice Romani 9, 16122 Genova, Italy\\
\email{damiano.verda@rulex.ai}}
\maketitle              
\begin{abstract}

In domains where transparency and trustworthiness are crucial, such as healthcare, rule-based systems are widely used and often preferred over black-box models for decision support systems due to their inherent interpretability. However, as rule-based models grow complex, discerning crucial features, understanding their interactions, and comparing feature contributions across different rule sets becomes challenging. To address this, we propose a comprehensive framework for estimating feature contributions in rule-based systems, introducing a graph-based feature visualisation strategy, a novel feature importance metric agnostic to rule-based predictors, and a distance metric for comparing rule sets based on feature contributions. By experimenting on two clinical datasets and four rule-based methods (decision trees, logic learning machines, association rules, and neural networks with rule extraction), we showcase our method's capability to uncover novel insights on the combined predictive value of clinical features, both at the dataset and class-specific levels. These insights can aid in identifying new risk factors, signature genes, and potential biomarkers, and determining the subset of patient information that should be prioritised to enhance diagnostic accuracy. Comparative analysis of the proposed feature importance score with state-of-the-art methods on 15 public benchmarks demonstrates competitive performance and superior robustness. The method implementation is available on GitHub: \url{https://github.com/ChristelSirocchi/rule-graph}.

\end{abstract}
\section{Introduction}

The integration of Machine Learning (ML) into various domains has revolutionised fields ranging from manufacturing to finance, offering unprecedented capabilities to learn from data and make accurate predictions. In safety- and ethic-critical domains such as healthcare, the ability to not only accurately predict outcomes but also to understand the decision-making process is crucial for ensuring transparency and trustworthiness~\cite{sirocchi2024medical}. Consequently, among various models and learning paradigms, a typical distinction is often drawn based on model interpretability into black-box and white-box models~\cite{sabbatini2021design}.
Black-box models achieve high accuracy by leveraging complex and often opaque relationships between inputs and outputs, while white-box models, such as rule-based systems, explicitly detail how inputs are transformed into outputs through predefined rules~\cite{masri2019survey}. These rules can be derived from domain knowledge or extracted from data using ML methods like decision trees and logic learning machines~\cite{rokach2005decision,lenatti2023characterization}, or derived from black-box models by rule extraction~\cite{sabbatini2021design}.

In high-stakes contexts such as the clinical domain, where maintaining control over the decision-making process is paramount, rule-based predictors remain prominent and preferred~\cite{pino2018gene}. However, as datasets accumulate and domain knowledge expands, these rule-based models can grow complex, often comprising hundreds of rules and patterns. Consequently, discerning which features and combinations thereof are most crucial for effective classification becomes challenging~\cite{solorio2020review}. To address this complexity, each rule-based ML model has developed its own feature importance strategy, such as Gini importance for decision trees and relevance scores for logic learning machines~\cite{nembrini2018revival,ferrari2023optimizing}. 
However, such metrics cannot explicitly identify which feature combinations and interactions are most relevant to the problem at hand, which may make the interpretation of the model harder for domain-expert users.
In this context, another challenge is to compare different rule sets, for instance, obtained from different models or over different datasets, as they typically vary in rule size and rule structure~\cite{katsuno1991difference,lenatti2023characterization}. Performing rules pairwise comparisons can quickly become computationally impractical for large rule sets and requires defining a measure of distance between rules to facilitate meaningful comparisons, which can be challenging~\cite{huysmans2007new,banaee2015data}.

Given the critical role of rule-based classifiers in high-stakes domains and the interpretability challenges they can present, we ask the following questions:

\begin{itemize}
    \item \emph{(a)} what features contribute the most to rules in rule-based classifiers?
    \item \emph{(b)} what feature combinations contribute to rules in these classifiers? 
    \item \emph{(c)} how similar/different are two rule sets over the same features with respect to the features' contribution to their rules?
\end{itemize}

\noindent
To tackle these questions, we propose a comprehensive and generalised framework for estimating feature contributions in rule-based systems which includes:

\begin{itemize}
    \item a graph-based feature visualisation strategy to explore relationships between features within rule sets across entire datasets or individual classes;
    \item a distance metric for comparing rule sets based on feature contributions, independent of rule size, facilitating comparisons across ML models and classes;
    \item a novel feature importance metric derived from the proposed feature graph that is agnostic to rule-based predictors and computationally efficient.
\end{itemize}

The proposed approach is extensively evaluated on both synthetic and benchmark datasets to assess its effectiveness in detecting relevant features and important feature interactions. A detailed analysis of two clinical datasets showcases the practical utility of our method, particularly in real-world biomedical applications. An implementation of the method is made available on GitHub for public access at \url{https://github.com/ChristelSirocchi/rule-graph}.


\section{Related work}

Rule-based systems are essential in high-stakes domains due to their inherent interpretability, which can become compromised by a large number of rules. 
A variety of visualisation strategies, distance metrics, and feature importance scores have been proposed to assess feature contribution in rule-based systems.
\Cref{sec:RBS} overviews algorithms for deriving rule sets, encompassing rule-based models learning rules from data and rule extraction methods approximating black-box models. 
\Cref{sec:RV} summarises available approaches for representing and visualising rules, \Cref{sec:RS} reviews existing strategies for comparing rule sets, while \Cref{sec:RI} examines feature importance and selection strategies.

\subsection{Rule-based systems}\label{sec:RBS}

Rule-based systems, originally termed expert systems due to their role in replacing or assisting human experts in knowledge-intensive tasks, offer distinct advantages by explicitly representing knowledge as rules: \emph{(a)} enabling domain experts to curate, update, and refine the rule set; \emph{(b)} deriving new knowledge through inference engines; and \emph{(c)} providing clear explanations to users for predictions~\cite{masri2019survey}.
Amidst the proliferation of black-box ML models, the transparency and explainability provided by rule-based systems have become particularly valuable. Consequently, rule-based systems remain widely used, especially in applications where consistency and transparency in decision-making are critical, such as expert systems for clinical decision support, assisting healthcare providers in applying clinical guidelines and treatment protocols~\cite{musen2021clinical}. Despite the improved predictive accuracy obtained by deep learning models, only a few are FDA-approved, and most predictive systems in healthcare remain rule-based.

Expert systems can leverage both expert-defined rules and rules learned from data. 
Decision Trees (DT) partition data recursively based on features, effectively handling mixed data types and non-linear relationships, though they risk over-fitting without proper pruning~\cite{rokach2005decision}.
Repeated Incremental Pruning to Produce Error Reduction (RIPPER), rooted in association rule mining, iteratively constructs rule sets by refining conditions to minimise classification errors, generating concise rule sets but struggling with noisy or irrelevant features~\cite{cohen1995fast}
Logic Learning Machine (LLM) uses shadow clustering and Boolean algebra to efficiently implement the switching neural network model and derive concise sets of rules~\cite{ferrari2023optimizing}. 
Fuzzy systems employ fuzzy logic to represent uncertainty and model imprecise data through degrees of membership in linguistic terms~\cite{pancho2013fingrams}.

To reconcile the accuracy of black-box models with the need for interpretability, symbolic knowledge extraction techniques have emerged, deriving interpretable rules from trained models. Provided that the extracted rules reflect the black-box behaviour with high fidelity, they can serve as an interpretable surrogate model or as a basis for constructing explanations.
Extraction algorithms are categorised into tree-based methods, like Classification and Regression Trees (CART), which recursively partition the feature space, and hypercube-based methods, such as ITER and GridEx, iteratively expanding in the input space~\cite{sabbatini2021design}.

\subsection{Rule sets representation and visualisation}\label{sec:RV}

To understand the relationship between rules and features in rule sets, various strategies have been proposed. Common rule set representations include lists and trees, with recent advancements introducing a layered graph structure, where the input layer represents input features, the conjunction layer the rule antecedents, and the output layer the class labels. Connections between input and conjunction layers denote condition judgements, while connections between conjunction and output layers signify mappings between rule antecedents and consequents~\cite{liu2016rule}.


Other graph-based visualisations are used in fuzzy systems, with rules as nodes and edges representing interactions between rules at the inference level in terms of rule co-firing~\cite{pancho2013fingrams}. In association rule mining, relationships between rules and features are represented as bipartite graphs (i.e. with nodes only sharing edges with nodes of the opposite type) and edge weights and node widths reflecting contributions computed as support, confidence, and lift~\cite{hahsler2017arulesviz}. However, these methods do not emphasise the relationships among features. 

Visualising feature interaction can be achieved by projecting the rule-feature bipartite graph onto the feature dimension. Available projection strategies are: \textit{(a) simple weighting}, with edges weighted by the frequency of common associations, \textit{(b) hyperbolic weighting}, addressing the decreasing marginal contribution of additional links, and \textit{(c) resource allocation}, which assumes each node has a certain amount of resources~\cite{zhou2007bipartite}. However, these strategies are defined for unweighted bipartite graphs and established weighted counterparts are still lacking. Moreover, these projections often cause unique relationships between nodes of different sets to disappear. In projecting rule sets onto feature graphs, accounting for edge weights connecting rules and features and preserving unique relationships is crucial, requiring a novel dedicated projection strategy.
\subsection{Rule sets distance methods}\label{sec:RS}

Evaluating the (dis)similarity of knowledge bases represented as rule sets plays a crucial role in several tasks: integrating diverse information sources, evaluating coherence with existing knowledge, ensuring backward compatibility during knowledge updates, detecting and eliminating duplicates or redundancies, and monitoring knowledge changes over time, highlighting evolving trends, emerging patterns, or shifts in underlying concepts~\cite{katsuno1991difference}. Additionally, comparing rule sets over different dataset splits enables assessing model consistency~\cite{lenatti2023characterization}.

Current methods for comparing rule sets typically involve two main approaches: one compares each rule in one set against all rules in another set, while the other matches each rule with the most similar rule in the other set based on criteria such as common rule outcomes~\cite{huysmans2007new} or pattern matching algorithms~\cite{banaee2015data}. These methods require defining a rule (dis)similarity criterion and often evaluate all possible pairwise rule comparisons, leading to substantial computational costs. Furthermore, these methods often assume that rules within each set are mutually exclusive, i.e. only one rule fires for each observation, which does not hold for some rule-based models like LLM.

Given these challenges, recent efforts have adopted a bag-of-words strategy to provide a vector representation for rules and used cosine similarity for rule set comparison, effectively overcoming computational challenges~\cite{lenatti2023characterization}. However, this approach does not quantify the contribution of features to rules, highlighting the need for more comprehensive methods for rule set comparison.




\subsection{Rule sets feature importance}\label{sec:RI}



Feature importance analysis plays a crucial role in improving model interpretability by pinpointing the most relevant input features and supporting feature selection efforts. 
In biomedical applications, feature importance strategies can enable clinicians to classify patients with distinct phenotypic characteristics using a small set of signature genes and potential biomarkers, facilitating the development of personalised screening tests and customised treatment~\cite{crippa2023characterization}. 

A model-agnostic strategy for evaluating feature contributions is permutation importance, which measures the reduction in model performance when the values of a feature are randomly shuffled \cite{altmann2010permutation}. However, it can be biased toward correlated features and can be sensitive to dataset size, permutation number, and chosen performance metric.
The field of XAI has introduced model-agnostic methods based on local explanations to uncover feature contributions to individual predictions. LIME (Local Interpretable Model-agnostic Explanations) perturbs the input data around a specific instance and fits a local interpretable model to approximate the black-box model locally, while SHAP (SHapley Additive exPlanations) uses Shapley values from cooperative game theory to quantify the contribution of each feature across all possible feature subsets~\cite{shap}.


In the context of rule-based predictors, dedicated strategies have been proposed.
For instance, features can be selected based on their presence or frequency in rule antecedents without considering their impact on rule effectiveness~\cite{chaves2012association}.
In fuzzy sets, the impact of each feature on fuzzy rules is evaluated using functions that estimate how well the feature predicts a class label~\cite{pino2018gene}.
Association rule mining relies on metrics like support, lift, and confidence to evaluate rule quality, primarily aimed at selecting rule subsets rather than analysing feature importance~\cite{tan2002selecting}.
Explicit feature importance scores are defined for LLM and DT. LLM computes the combined relevance of each condition containing that feature~\cite{lenatti2023characterization} while DTs use information gain and entropy-based methods, such as Gini importance, to calculate the reduction in impurity or entropy brought by a feature across all nodes that use it to split the data~\cite{nembrini2018revival}.

Additional methods for identifying relevant features have been developed in the field of feature selection, categorised based on their relationship with the predictor. \textit{Filter methods} use statistical metrics such as correlation or variance and, while computationally efficient, do not offer insights into how the model uses features. In contrast, \textit{wrapper methods} search for a feature subset that optimises a predefined criterion when the algorithm is trained on this subset and pose challenges such as the need for an appropriate criterion and high computational costs~\cite{solorio2020review}. Notably, since feature selection is often treated as a step-wise process, these methods often fail to exploit the interaction between features.





\section{Proposed approach}
Starting from a layered rule representation akin to Liu et al.~\cite{liu2015network}, we consider three sets of nodes: features, rules, and classes. In our approach, connections between feature nodes and rule nodes are weighted edges, indicating the contribution of each feature to the corresponding rule (\emph{feature relevance}). Similarly, connections between rule nodes and class labels are weighted edges, representing the contribution of each rule to class prediction (\emph{rule relevance}). We propose a projection strategy to map this tripartite graph onto the feature set such that edges between features reflect their shared contribution to the same rules, and the centrality of each feature reflects its overall importance across all rules and serves as a feature importance metric. We extend this strategy to construct class-specific feature graphs and define a distance metric to compare graphs.



\paragraph{\textbf{Rule set.}}
Let $\mathcal{D}$ represent a dataset comprising $d$ samples denoted by $\boldsymbol{x}_s$, with $s$ from 1 to $d$. Each sample is described by $m$ input features, defined over the feature set $\mathcal{V} = \{v_1, v_2, \ldots, v_m\}$. 
For each input $\boldsymbol{x}_s$, $y_s$ denotes the corresponding target. In classification tasks, the target takes discrete values in $\mathcal{T} = \{ t_1, t_2, \ldots, t_r \}$.
Then, a rule set $\mathcal{R}$ can be defined over $\mathcal{D}$, mapping instances to targets, and consists of a set of rules each denoted by $R$. If $\mathcal{R}$ has $n$ rules, then $\mathcal{R} = \{ R^1, R^2, \ldots, R^n \}$.
Each $R$ is a logical expression where the antecedent or premise of the rule is a set of conditions over the features of the dataset, while the consequent is the outcome (here class assignment) when all conditions specified by the antecedent are met. Formally, a rule $R^k$ can be denoted as a pair $R^k = (C^k, T^k)$, with $C^k = \{c_1^k, c_2^k, \ldots, c_q^k\}$ and $T^k \in \mathcal{T}$, where $C^k$ denotes the set of $q$ conditions in the rule and $T^k$ is the target associated with that rule:
$$
c_1^k \land c_2^k \land \cdots \land c_q^k \implies T^k
$$
Let $\mathcal{I}_{\square}$ be a function that computes the relevance of a feature $v$ for a rule $R$ in a dataset $\mathcal{D}$. For a rule set $\mathcal{R}$, a feature relevance matrix $\boldsymbol{P}$ can be defined as the $n \times m$ matrix, with $n$ the number of rules in $\mathcal{R}$ and $m$ the number of features, of the elements $[p_{ij}]$, where
\begin{equation}\label{eq:fr}
p_{ij} = \mathcal{I}_{\square}(\mathcal{D}, v_i, R^j) \quad \forall v_i\in \mathcal{V},\; \forall R^j \in \mathcal{R}.
\end{equation}
Additionally, let $\mathcal{I}_{\nabla}$ be a function that computes the relevance of a rule $R$ in a rule set $\mathcal{R}$ over a dataset $\mathcal{D}$. For a rule set $\mathcal{R}$, a rule relevance vector $\boldsymbol{q}$ of length $n$ can be defined with elements $q_{j}$, where
\begin{equation}\label{eq:rr}
    q_j = \mathcal{I}_{\nabla}(\mathcal{D}, R^j) \quad \forall R^j \in \mathcal{R}. 
\end{equation}

\paragraph{\textbf{Graph projection \& visualisation.}} 
Let $\boldsymbol{A}$ be a $m \times m$ matrix, with $m$ the number of features, of the elements $[a_{ij}]$ such that
\begin{equation}\label{eq:proj}
 a_{ij} = 1 - \prod_{k = 1}^{n} (1 - p_{ki} \cdot p_{kj} \cdot q_k) \quad \forall i, j \in \{1, \ldots, m\}   
\end{equation}
$\boldsymbol{A}$ is then normalised such that the sum of all its elements is 100, to obtain $\boldsymbol{A}^\prime$:
$$
A_{ij}^\prime = \frac{A_{ij}}{\sum_{i,j=1}^{m} A_{ij}} \cdot 100 \quad \forall i, j \in \{1, \ldots, m\}
$$
$\boldsymbol{A}^\prime$ is the adjacency matrix of a weighted and undirected feature graph, which can be visualised to examine the feature interactions within the rule set $\mathcal{R}$ over the dataset $\mathcal{D}$.

According to the proposed projection strategy, the product $p_{ki} \cdot p_{kj} \cdot q_k$ in \Cref{eq:proj} captures the joint relevance of features $v_i$ and $v_j$ with respect to rule $R^k$, scaled by $q_k$ to also account for the relevance of the rule. These contributions are aggregated across all rules in a multiplicative manner, making the overall score more sensitive to instances where two features exhibit high joint relevance in at least one relevant rule, in contrast to simple summation. In fact, in rule sets, it is not expected that two features interact as strongly across all rules, and strong interactions can be noteworthy, even if infrequent.
Moreover, this projection strategy generates self-edges $a_{ii}$ for each feature $i$ in $\{1, \dots, m\}$, quantifying its individual contribution across all rules. These self-edges also account for instances where a feature appears alone in a rule, which are often crucial to the rule set but are lost in most projection strategies. 

\paragraph{\textbf{Class-specific graph projection.}}
A feature graph specific for a given class $t \in \mathcal{T}$ can be constructed by considering only rules having the given class as consequent. 
Let $\mathcal{R}_i$ be the subset of rules in $\mathcal{R}$ with target $t_i$, i.e. $\mathcal{R}_i = \{R^k \mid R^k \in \mathcal{R} \land T^k = t_i\}$. $\boldsymbol{A}_i^\prime$ is the adjacency matrix of the feature graph defined as in \Cref{eq:proj} but where the matrix $\boldsymbol{P}$ and the vector $\boldsymbol{q}$ are defined over the rule set $\mathcal{R}_i$ rather than $\mathcal{R}$.

\paragraph{\textbf{Graph distance.}}
The distance between two graph representations can be computed as the distance between the respective adjacency matrices $\boldsymbol{A}_1^\prime$ and $\boldsymbol{A}_2^\prime$:
\begin{equation}
d(\boldsymbol{A}_1^\prime, \boldsymbol{A}_2^\prime) = \|\boldsymbol{A}_1^\prime - \boldsymbol{A}_2^\prime\|_F
\end{equation}
where the Frobenius norm $\|\boldsymbol{A}\|_F$ of a matrix $\boldsymbol{A}$ is given by:
$$
\|\boldsymbol{A}\|_F = \sqrt{\sum_{i,j=1}^{n} |A_{ij}|^2}
$$

\paragraph{\textbf{Feature importance.}}
A feature importance score can be computed for features in $\mathcal{V}$ as the degree centrality of the nodes of the graph defined by $\boldsymbol{A}^\prime$. Specifically, the importance of $v_i$ is given by the sum of the elements in the $i$-th row of $\boldsymbol{A}^\prime$:
\begin{equation}
    \text{Importance}(v_i) = \sum_{j=1}^{m} A_{ij}^\prime
\end{equation}
This metric aggregates contributions from both self-edges and edges with other features, capturing both the independent and combined feature contributions.

\paragraph{\textbf{Relevance metrics.}}
Feature and rule relevance metrics, proposed in LLM and adopted in this study, leverage the concepts of \textit{error} and \textit{covering} based on the fraction of data samples assigned to a class and satisfying a rule.

A data sample $ \boldsymbol{x}_s $ satisfies a rule $ R^k $ if all its conditions are true for $ \boldsymbol{x}_s $, i.e.,
$$
\boldsymbol{x}_s \models R^k \quad \iff \quad c_1^k(\boldsymbol{x}_s) \land c_2^k(\boldsymbol{x}_s) \land \ldots \land c_q^k(\boldsymbol{x}_s)
$$
where $ c_i^k(\boldsymbol{x}_s) $ denotes the evaluation of condition $ c_i^k $ on sample $ \boldsymbol{x}_s $.
The subset $ \mathcal{D}^k $ of $ \mathcal{D} $ satisfying the conditions $ C^k $ of $ R^k $ can be defined as:
$$
\mathcal{D}^k = \{ \boldsymbol{x}_s \in \mathcal{D} \mid \boldsymbol{x}_s \models R^k \}
$$
Additionally, let $ \mathcal{D}_i $ represent the subset of all samples in $ \mathcal{D} $ that are assigned to a given target class $ t_i $, and let $ \mathcal{D}_i^\prime $ be the subset of samples not assigned to $ t_i $:
$$
\mathcal{D}_i = \{ \boldsymbol{x}_s \in \mathcal{D} \mid y_s = t_i \} \quad \mathcal{D}_i^\prime = \{ \boldsymbol{x}_s \in \mathcal{D} \mid y_s \neq t_i \}
$$
Then, \textit{covering} can be defined as the proportion of samples assigned to the correct class $T^k = t_i$ that satisfy the rule $R^k$, while \textit{error} is the proportion of samples assigned to a different class that satisfy the rule $R^k$, i.e.,
\begin{equation}
\textit{covering}(R^k) = \frac{|\mathcal{D}^k \cap \mathcal{D}_i|}{|\mathcal{D}_i|}
\quad 
\textit{error}(R^k) = \frac{|\mathcal{D}^k \cap \mathcal{D}_i^\prime|}{|\mathcal{D}_i^\prime|}
\end{equation}
and \textit{rule relevance} can be defined as:
\begin{equation}
\mathcal{I}_{\nabla}(\mathcal{D}, R^k) = \text{covering}(R^k) \cdot (1 - \text{error}(R^k))
\end{equation}
Moreover, adapting from~\cite{ferrari2023optimizing}, let $R_{-h}^{k}$ be the rule obtained from $R^{k}$ by removing the conditions on feature $v_h$, i.e.,
$$
R_{-h}^{k} = (C_{-h}^{k}, T^k), \quad C_{-h}^{k} = \{c^k \mid c^k \in C^{k} \land V(c^k) \neq v_h\}
$$
where $V(c^k)$ denotes the feature over which the condition is applied. 
Then, \textit{feature relevance} can be computed as the increase in error to the rule as a result of removing the conditions over the feature:
\begin{equation}
\mathcal{I}_{\square}(\mathcal{D}, v_h, R^k) = (\text{error}(R_{-h}^k) - \text{error}(R^k)) \cdot \text{covering}(R^k)
\end{equation}
Other rule relevance metrics include support, confidence, and lift, while an alternative feature relevance score is impurity gain (all implemented in our code).






\section{Evaluation strategy}

\subsection{Feature graph evaluation on synthetic dataset}
The proposed approach was first evaluated on synthetic datasets to showcase the advantages of graph feature representation with respect to feature importance scores, particularly in terms of the ability to differentiate between features that are predictive of the target independently or combined. Three types of datasets were generated: (1) with all relevant features independently predictive of the target, (2) with all relevant features predictive to the target when combined, and (3) with a mix of independently and combined predictive features. Each dataset comprised 2000 instances and 8 features, with relevant features varying from 2 to 6, and 10 datasets were generated per configuration using different random seeds. All features were uniformly sampled in the range [0,1].
%
Independent predictive features were obtained by assigning a set of intervals in the range [0,1] to each feature. For each data sample, the target was set to 1 if the value of at least one feature fell within its corresponding predefined interval.
In contrast, combined predictive features were generated by setting the target variable to 1 if the majority of the feature values in the data sample exceeded a given threshold.
These two strategies were also combined to obtain datasets with some independent and some combined relevant features. For further implementation details, refer to the GitHub repository for reproducing the synthetic datasets.
%

%
The rule-based classifier adopted in this evaluation was DT, with experiments also conducted using LLM, yielding similar results (not shown). DTs were trained using 5-fold nested cross-validation with hyperparameter tuning and converted to rule sets by translating each path from the root to the leaves into if-then rules. Feature graphs were constructed from these rule sets using the proposed method, with adjacency matrices visualised as heatmaps. Gini importance was also computed, serving as a standard feature importance metric for comparison.

\subsection{Feature graph evaluation on benchmark datasets}
The potential of the proposed approach to identify relevant features and their interactions was evaluated on public datasets, primarily sourced from the clinical domain, the main application target of this method. To ensure broader applicability, datasets from other domains were also included. The selected datasets were diverse in the number of instances (from 100 to 5000), number of features (from 4 to 240), number of classes (from 2 to 10), and feature types (categorical, continuous, and mixed). 
Two datasets were analysed in depth.
The Pima Indians Diabetes dataset\footnote{\url{https://www.kaggle.com/datasets/uciml/pima-indians-diabetes-database}}, with 768 medical profiles and 8 clinical features for diabetes detection, was used as a binary classification case study to examine and compare feature contributions across models.
The Breast Tissue dataset\footnote{\url{https://archive.ics.uci.edu/dataset/192/breast+tissue}}, comprising 106 instances with 9 attributes of electrical impedance measurements of freshly excised breast tissue samples, was used for a multi-class classification case study to detect class-specific feature interactions.
For these datasets, rule sets were generated according to four strategies: LLM\footnote{The Rulex platform, www.rulex.ai, was used for its implementation of the LLM}, DT, RIPPER, and a MultiLayer Perceptron (MLP) with rule extraction using a CART implementation available from the PsyKE library~\cite{sabbatini2021design}. The fidelity of the extracted rule sets was evaluated by computing accuracy and F1-score compared to the black-box, and the number of extracted rules was set to 15 to balance readability and fidelity.


\subsection{Feature selection performance and robustness on benchmarks}
The effectiveness of feature centrality as a measure of feature importance, assessed by its ability to identify the top-$k$ features in a dataset, was evaluated across 15 benchmark datasets. Our graph-based importance score was compared against three feature importance metrics: permutation importance, Gini importance, and average SHAP values. DTs were trained using 5-fold nested cross-validation to derive rule sets and generate feature graphs.
Evaluation criteria encompassed both performance and robustness of the feature importance scores. Performance evaluation involved selecting the top 5 and top 10 features identified by each metric, training decision trees on these features, and computing prediction accuracy. Robustness was assessed by calculating the average pairwise Spearman's correlation across cross-validation folds for each importance score.
%



\section{Results}

\subsection{Feature graph evaluation on synthetic dataset}

\begin{figure}[t!]
\centering 
\includegraphics[width=\textwidth]{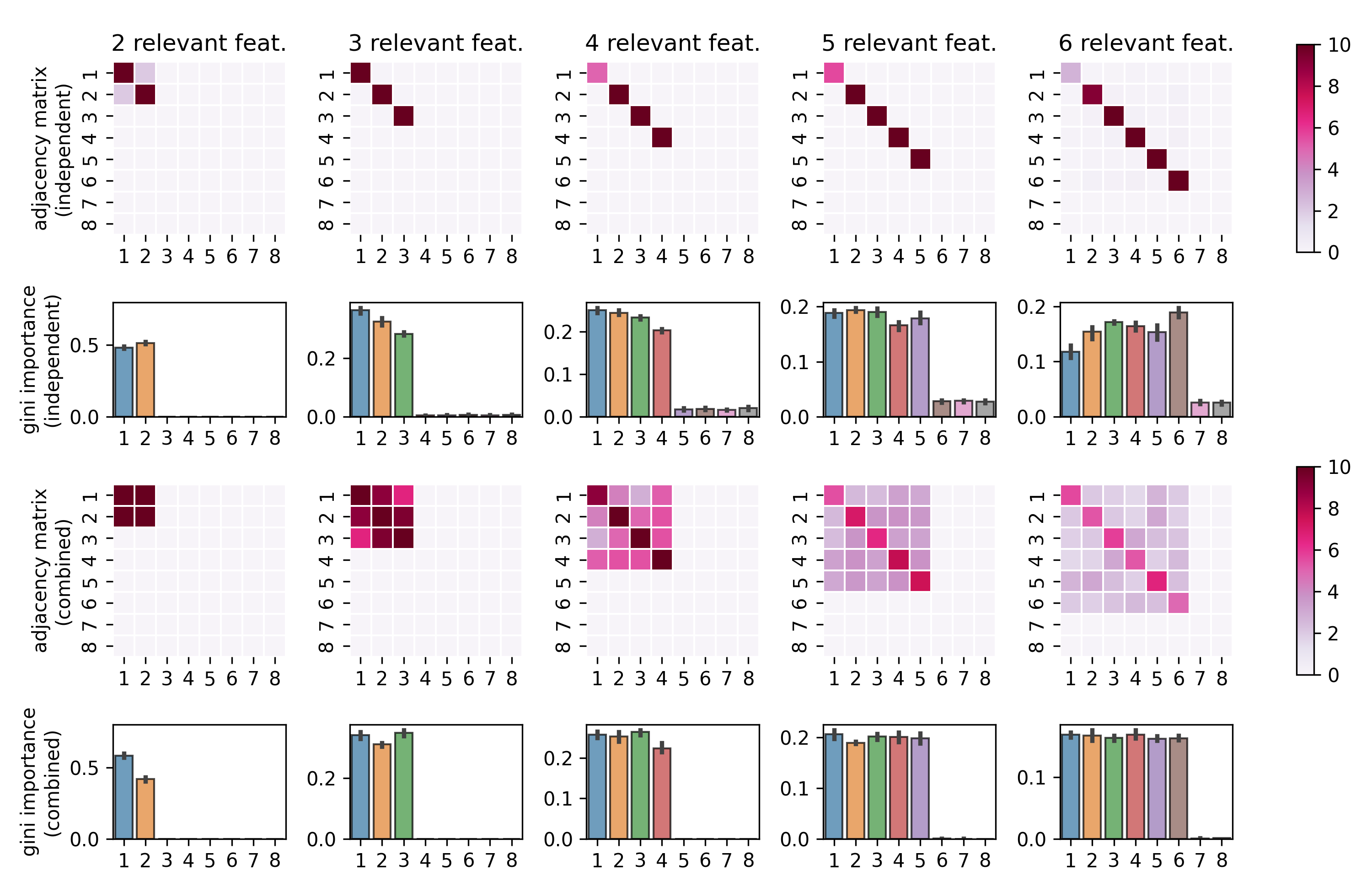}
\caption{Adjacency matrices constructed according to the proposed approach and Gini importance indices obtained from decision trees trained on synthetic datasets comprising a number of relevant features ranging from 2 to 6, which are predictive of the target class either independently or combined.
}
\label{fig:synthetic1}
\end{figure}

The evaluation of the proposed approach on synthetic datasets underscores the superiority of a graph-based feature representation over a one-dimensional feature importance score for uncovering the collective roles of features in class prediction. \Cref{fig:synthetic1} presents the average adjacency matrix of feature graphs built on rule sets derived for synthetic datasets with varying numbers of relevant features and feature configurations. For independently predictive features (first row), the heatmap reveals heavy weights on the diagonal (self-edges), indicating that each rule primarily relies on a single highly predictive variable. Connections between different features show weaker weights, suggesting minimal influence from other variables. In contrast, combined predictive features (third row) exhibit heavier weights on edges connecting different features, indicating collaborative predictive power among multiple features.
While Gini importance effectively distinguishes relevant from irrelevant features, it fails to differentiate between features that are predictive independently or combined. 
Intermediate scenarios, comprising a mix of these two types of features, were also explored, yielding similar insights.
These results underscore the advantage of our feature graph approach, providing a more nuanced representation of feature interactions.


\subsection{Feature graph evaluation on benchmark datasets}

\begin{figure}[!t]
\centering 
\includegraphics[width=\textwidth]{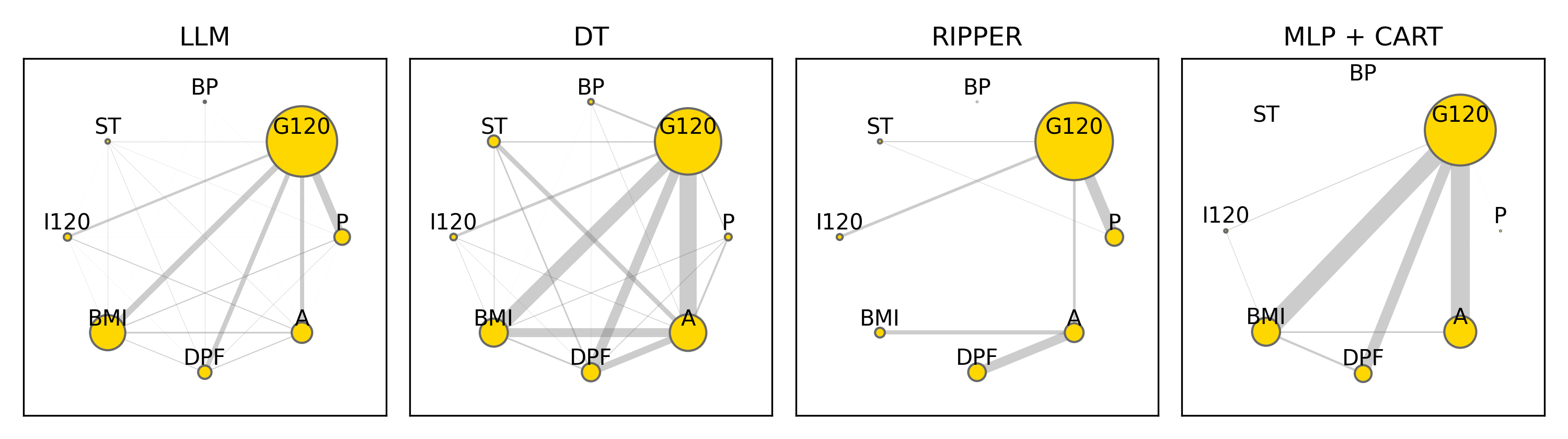}
\caption{Evaluation on the Pima Indians Diabetes dataset, comprising 8 features: pregnancies (P), glucose at 2 hours (G120), blood pressure (BP), skin thickness (ST), insulin at 2 hours (I120), body mass index (BMI), diabetes pedigree function (DPF), and age (A). Four rule sets were generated using LLM, DT, RIPPER, and MLP with rule extraction via CART. Feature graphs are visualised with node size proportional to its centrality (self-edges not shown). 
}
\label{fig:pima}
\end{figure}

\textbf{Rule set visualisation. }
Feature graphs constructed from rule sets over the Pima Indians Diabetes dataset illustrate the valuable insights this representation can provide on feature interactions in a clinical setting.
\Cref{fig:pima} displays feature graphs corresponding to rule sets obtained from four learning schemes (LLM, DT, RIPPER, MLP + CART), where node size is proportional to its centrality (our defined feature importance score).
Node centrality reveals that all models identify glucose levels (G120) as the most crucial feature for diabetes classification, which aligns with the clinical understanding that diabetes is characterised by elevated blood sugar. Age is identified as the second most important risk factor by DT and MLP and the third by LLM and RIPPER. Body mass index (BMI) and family history of diabetes (DPF) are also consistently leveraged across all models, confirming known diabetes risk factors. In fact, according to the NIH, the top three risk factors for diabetes are being overweight or obese, being 35 years or older, and having a family history of diabetes\footnote{\url{https://www.niddk.nih.gov/health-information/diabetes/overview/risk-factors-type-2-diabetes}}. The number of pregnancies is considered by LLM and RIPPER, reflecting the role of gestational diabetes.
Other features, which have positive correlations with the target and appear in the rule sets, show weak or null edge weights and very low centrality. This indicates that these features do not substantially contribute to the rules and their removal does not strongly affect the model predictive ability.

Feature interactions reveal a heavy edge between Glucose and Age in DT, and MLP, indicating that age is predictive of diabetes when glucose levels are also elevated. Similarly, a strong edge between BMI and Glucose suggests that BMI's predictive power is enhanced when considered alongside glucose levels. This aligns with established protocols that predict high diabetes risk when both features are elevated~\cite{kunapuli2010online,sirocchi2024medical}. Additionally, LLM and RIPPER identify interactions between Glucose and the number of pregnancies. Minor interactions not involving Glucose are found in DT and RIPPER between DPF and Age, and between BMI and Age, suggesting that the impact of BMI on diabetes risk varies with age and reflecting the established association between obesity and age as risk factors for diabetes~\cite{luo2020age}.
Overall, the analysis shows that Glucose is most predictive, while other factors (BMI, Age, DPF, pregnancies) can be predictive, but primarily in combination with Glucose. Understanding these combined predictive values of clinical features is crucial for determining which patient information should be collected and prioritised to enhance diagnostic accuracy.\\

\noindent
\textbf{Rule sets comparison.}
The pair-wise distance calculated between the adjacency matrices of the feature graphs in \Cref{fig:pima} shows that the two most similar models (in terms of feature contribution to extracted rules) are DT and MLP, whose graphs are dominated by Glucose-BMI and Glucose-Age interactions. In contrast, the most different models are DT and RIPPER, which differ in the Glucose-Age, and Glucose-BMI interactions (stronger in DT) and Glucose-pregnancies (stronger in RIPPER). 
Evaluating the distance between rule sets, facilitated by our graph representation, can be beneficial in various scenarios, particularly for ensuring consistency and continuity in decision-making. When updating a rule-based expert system with a new, more predictive rule set, selecting the one with the most similar feature graph to the existing system ensures that predictions and explanations remain consistent. In clinical settings, this supports the concept of continuity of care, defined as the consistency of healthcare events experienced by individuals over time and across different providers~\cite{haggerty2003continuity}.

The pair-wise distance calculated between feature graphs constructed from DTs of varying depths trained on the Pima dataset demonstrated high similarity, suggesting robustness against over-fitting. These findings confirm that the proposed approach, by evaluating a feature contribution in terms of the quality of rules with that feature removed, effectively handles scenarios where less relevant features are included, arising when a model grows complex and over-fits the data, thus reducing the need for extensive hyperparameter tuning, unfeasible in some resource-constrained scenarios.
Moreover, in this experiment, feature centrality rankings remained more stable across different tree depths compared to Gini importance, with an average Spearman's correlation of 0.98 versus 0.94, confirming the greater ranking robustness offered by our approach.\\

\begin{figure}[!t]
\centering 
\begin{subfigure}[b]{1\textwidth}
    \centering
    \includegraphics[width=\textwidth]{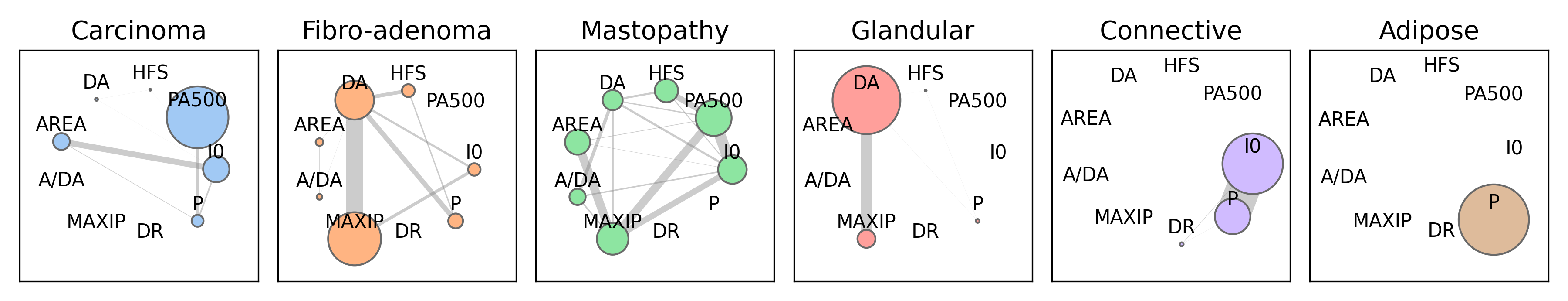}
    \caption{Feature graphs from decision tree}
\end{subfigure}

\begin{subfigure}[b]{1\textwidth}
    \centering
    \includegraphics[width=\textwidth]{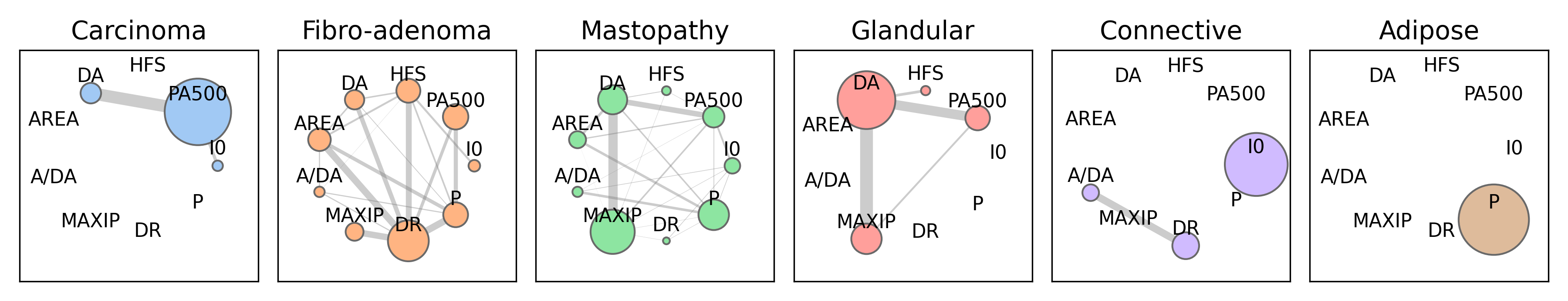}
    \caption{Feature graphs from logic learning machine}
\end{subfigure}

\caption{Evaluation on the Breast Cancer dataset, comprising 9 breast tissue characteristics: Impedivity (I0), phase angle at 500 KHz (PA500), high-frequency slope of phase angle (HFS), impedance distance between spectral ends (DA), area under spectrum (AREA), area normalised by DA (A/DA), maximum of the spectrum (MAXIP), distance between I0 and real part of the maximum frequency point (DR), and spectral curve length (P).  Class-specific feature graphs were constructed on rule sets obtained from DT and LLM (self-edges not shown).}
\label{fig:class_graph}
\end{figure}

\noindent
\textbf{Class-specific rule set visualisation.}
Class-specific feature graphs were generated for the Breast from rule sets derived from DT and LLM (\Cref{fig:class_graph}).
For both models, predicting fibroadenoma and mastopathy proved the most challenging, requiring nearly all features and exhibiting multiple feature interactions. This is unsurprising, as these two tissues can be hardly distinguished even in histopathology and are often grouped under the same class.
In contrast, predicting adipose tissue relied solely on the length of the spectral curve (P), without any feature interactions. In fact, this feature indicates the overall complexity and variability of the tissue's impedance profile, distinguishing homogeneous tissues like adipose from others with more complex impedance spectra.
For predicting carcinoma, the phase angle at 500 KHz (PA500) emerged as the most important feature. Indeed, carcinomas are characterised by higher cellularity and different extracellular matrix compositions and exhibit distinct phase angles. This finding aligns with literature linking phase angle to survival in advanced cancer patients \cite{hui2017association}. Both models identified interactions between PA500 and impedance (DA) and/or impedivity (I0), while DT detected an interaction between impedivity and the area under the impedance spectrum (AREA), all features that provide further information on the tissue's electrical properties. 
In predicting glandular tissue, both models utilised the impedance (DA) - Max spectrum (MAX IP) feature combination, with LLM also incorporating phase angle. In connective tissue prediction, impedivity was crucial, either independently (in LLM) or in conjunction with the spectral curve length (in DT).
Overall, the two models detected similar interactions, with variations likely due to the different learning paradigms, the scarcity of data samples, and the redundant nature of features. Nonetheless, these explorations offer valuable insights into what sets of measurements should be collected to distinguish a given diagnostic group from others.


\subsection{Feature selection performance and robustness on benchmarks}

Classification accuracy results for 15 benchmark datasets (\Cref{table:performance}) show that when selecting the top 5 features, both our graph-based importance score and permutation importance were highly effective, achieving 7 and 8 top accuracy scores, respectively. In contrast, Gini importance excelled only once, while SHAP never did. For the top 10 features, centrality and permutation importance again performed best, with 8 and 7 top scores, respectively, while Gini and SHAP each excelled in only two datasets. Overall, the proposed approach performs comparably to permutation importance, a well-established metric, and surpasses Gini and SHAP.
Evaluation of feature rank robustness demonstrates that the proposed approach is the most stable, achieving 8 top scores across 15 datasets. Gini importance follows with 4 top scores, SHAP with 2, and permutation with 1, indicating its sensitivity to data splits. Therefore, while the proposed importance metric performs comparably to permutation in accuracy, it excels in robustness, making it the most reliable metric across the evaluated benchmarks by effectively mitigating the impact of data variability.

\begin{table}[!t]
\centering
\centerline{
\begin{tabular}{|c|cccc|cccc|cccc|}
\toprule
                       & \multicolumn{4}{|c|}{Accuracy Top 5 features} & \multicolumn{4}{c|}{Accuracy Top 10 features} & \multicolumn{4}{c|}{Importance Robustness} \\ \midrule
                       & Perm & Gini & SHAP & \underline{Graph} & Perm & Gini & SHAP & \underline{Graph} & Perm & Gini & SHAP & \underline{Graph} \\ \midrule
\textbf{Hill Valley}   & 0.541 & 0.527 & 0.540 & \textbf{0.554} & 0.535 & 0.541 & \textbf{0.554} & 0.551 & 0.114 & 0.246 & \textbf{0.251} & 0.247 \\
\textbf{Hypothyroid}   & \textbf{0.979} & 0.974 & 0.975 & 0.975 & \textbf{0.977} & \textbf{0.977} & 0.974 & \textbf{0.977} & 0.656 & 0.892 & 0.889 & \textbf{0.892} \\
\textbf{Pixel}         & 0.634 & 0.640 & 0.633 & \textbf{0.650} & 0.807 & \textbf{0.852} & 0.836 & 0.826 & \textbf{0.374} & 0.338 & 0.334 & 0.336 \\
\textbf{Tokyo}         & 0.916 & 0.913 & 0.911 & \textbf{0.916} & \textbf{0.912} & 0.899 & 0.902 & 0.904 & 0.315 & 0.424 & 0.414 & \textbf{0.450} \\
\textbf{Balance Scale} & \textbf{0.789} & \textbf{0.789} & 0.787 & \textbf{0.789} & \textbf{0.789} & \textbf{0.789} & 0.787 & \textbf{0.789} & 0.076 & \textbf{0.112} & 0.048 & 0.068 \\
\textbf{Breast Cancer} & 0.693 & 0.724 & 0.731 & \textbf{0.735} & 0.697 & \textbf{0.725} & 0.707 & 0.711 & 0.133 & 0.383 & 0.398 & \textbf{0.405} \\
\textbf{BCW Original}  & \textbf{0.950} & 0.944 & 0.944 & 0.947 & 0.941 & \textbf{0.944} & 0.941 & \textbf{0.944} & 0.637 & \textbf{0.757} & 0.725 & 0.752 \\
\textbf{BCW Prognostic}& \textbf{0.737} & 0.727 & 0.722 & 0.718 & 0.717 & 0.707 & 0.712 & \textbf{0.727} & 0.098 & 0.403 & 0.397 & \textbf{0.421} \\
\textbf{BCW Diagnostic}& 0.935 & \textbf{0.937} & 0.930 & 0.928 & 0.924 & 0.923 & \textbf{0.928} & 0.921 & 0.314 & \textbf{0.375} & 0.361 & 0.360 \\
\textbf{Car Evaluation}& 0.819 & 0.815 & 0.823 & \textbf{0.825} & 0.924 & 0.922 & 0.919 & \textbf{0.926} & 0.794 & \textbf{0.814} & 0.795 & 0.742 \\
\textbf{Contraceptive} & \textbf{0.561} & 0.558 & 0.558 & 0.559 & 0.558 & \textbf{0.559} & \textbf{0.559} & \textbf{0.559} & 0.844 & 0.826 & 0.844 & \textbf{0.847} \\
\textbf{Hepatitis}     & 0.784 & 0.772 & 0.741 & \textbf{0.785} & \textbf{0.818} & 0.779 & 0.756 & 0.779 & 0.364 & 0.687 & 0.634 & \textbf{0.710} \\
\textbf{SPECT Heart}   & \textbf{0.791} & 0.783 & 0.783 & 0.772 & \textbf{0.768} & 0.767 & 0.753 & \textbf{0.768} & 0.128 & 0.517 & 0.516 & \textbf{0.524} \\
\textbf{Waveform} & \textbf{0.744} & 0.740 & 0.741 & 0.740 & \textbf{0.775} & 0.772 & 0.772 & \textbf{0.775} & 0.802 & 0.911 & \textbf{0.909} & 0.869 \\
\textbf{Zoo}           & \textbf{0.931} & 0.880 & 0.900 & 0.881 & \textbf{0.950} & 0.930 & 0.930 & 0.940 & 0.701 & 0.751 & 0.757 & \textbf{0.758} \\
\midrule
\end{tabular}
}
\caption{Accuracy scores for DTs trained on the top 5 and 10 features identified by permutation Gini importance, average SHAP values, and our approach, as well as robustness of feature ranks for each method, with the best values in bold.
}
\label{table:performance}
\end{table}

\section{Conclusion and future work}
Our proposed framework effectively enhances the interpretability of rule-based systems by estimating feature contributions using a graph-based visualisation strategy, a novel feature importance metric, and a distance metric for comparing rule sets. Extensive evaluation on synthetic and benchmark datasets demonstrates the framework's ability to uncover crucial features and interactions, providing valuable insights for high-stakes domains such as healthcare. 
Future work will explore several avenues to enhance the current methodology. 
First, we plan to investigate additional rule-based predictors and hypercube-based rule extraction methods. 
Second, we aim to evaluate additional rule and feature relevance metrics such as support, confidence, lift, and impurity gain. 
Lastly, we will explore alternative graph construction strategies, including modelling relationships as hypergraphs, as well as diverse centrality measures and graph mining methods to enhance the visualisation and interpretation of feature interactions.




\bibliographystyle{splncs04}
\bibliography{bibliography}




\end{document}